\pdfoutput=1
\documentclass{ifacconf}

\usepackage{graphicx}      % include this line if your document contains figures

\usepackage[numbers]{natbib}
\usepackage{textcomp}
\usepackage{stfloats}
\usepackage{url}
\usepackage{verbatim}
\usepackage{graphicx}
\usepackage{booktabs}
\usepackage{amsmath,amsfonts}

\usepackage{etoolbox}
\let\classAND\AND
\let\AND\relax
\usepackage{algorithmic}
\let\AND\classAND
\AtBeginEnvironment{algorithmic}{\let\AND\algoAND}

\usepackage{algorithm}

\usepackage{array}
\usepackage{subcaption}

\makeatletter
\let\old@ssect\@ssect % Store how ifacconf defines \@ssect
\makeatother

\usepackage{natbib}
\usepackage{hyperref}
\hypersetup{
    colorlinks=true,
    linkcolor=green,
    citecolor=green,
    filecolor=green,
    urlcolor=green,
}
\makeatletter
\def\@ssect#1#2#3#4#5#6{%
  \NR@gettitle{#6}% Insert key \nameref title grab
  \old@ssect{#1}{#2}{#3}{#4}{#5}{#6}% Restore ifacconf's \@ssect
}
\makeatother

\begin{document}
\begin{frontmatter}

\title{Learning Heterogeneous Agent Cooperation via Multiagent League Training
\thanksref{footnoteinfo}}

\thanks[footnoteinfo]{This work was supported in part by 
the National Key Research and Development Program of China (2018AAA0102404), 
the National Natural Science Foundation of China (62073323), 
the Strategic Priority Research Program of Chinese Academy of Sciences (XDA27030204),
the External Cooperation Key Project of Chinese Academy Sciences (173211KYSB20200002),
the Science and Technology Development Fund of Macau (No.0025/2019/AKP),
and 
the Beijing Nova Program under Grant 20220484077.}


% \thanks[footnoteinfo]{Sponsor and financial support acknowledgment
% goes here. Paper titles should be written in uppercase and lowercase
% letters, not all uppercase.}

\author[1,2]{Qingxu Fu} 
\author[1,2]{Xiaolin Ai} 
\author[1,2]{Jianqiang Yi}
\author[1,2]{Tenghai Qiu} 
\author[3]{Wanmai Yuan}
\author[1,2]{Zhiqiang Pu} 

\address[1]{\small{Institute of Automation, Chinese Academy of Sciences. }}   % . Beijing, China.
\address[2]{\small{School of Artificial Intelligence, University of Chinese Academy of Sciences.}} %  of Sciences. Beijing, China.
\address[3]{\small{Electronics Technology Group Corporation, Information Science Academy of China. Beijing, China.}}

\begin{abstract}                % Abstract of not more than 250 words.
   Many multiagent systems in the real world include multiple types of agents with different abilities and functionality. Such heterogeneous multiagent systems have significant practical advantages. However, they also come with challenges compared with homogeneous systems for multiagent reinforcement learning, such as the non-stationary problem and the policy version iteration issue. This work proposes a general-purpose reinforcement learning algorithm named Heterogeneous League Training (HLT) to address heterogeneous multiagent problems. HLT keeps track of a pool of policies that agents have explored during training, gathering a league of heterogeneous policies to facilitate future policy optimization. Moreover, a hyper-network is introduced to increase the diversity of agent behaviors when collaborating with teammates having different levels of cooperation skills. We use heterogeneous benchmark tasks to demonstrate that (1) HLT promotes the success rate in cooperative heterogeneous tasks; (2) HLT is an effective approach to solving the policy version iteration problem; (3) HLT provides a practical way to assess the difficulty of learning each role in a heterogeneous team. 
\end{abstract}


\end{frontmatter}
%===============================================================================
\section{Introduction}
% 缺少通用性HMARL算法 
% \IEEEPARstart{H}{eterogeneous}
Heterogeneous Multiagent Reinforcement Learning (HMARL) is not a new problem in MARL,
but they have not yet been investigated extensively in the literature.
A number of studies have covered HMARL, investigating its application in specific domains
such as communication \cite{fard2022time} and UAV allocation \cite{zhao2019fast}.
However, to the best of our knowledge, 
there are currently no general-purpose heterogeneous reinforcement learning algorithms 
for cooperative learning within a multi-type, multiagent team.

% 一般MARL算法, 异构性不足
In general, classic MARL algorithms do not assume the type difference between agents.
Nevertheless,
most existing research on MARL is primarily focused on
homogeneous environments \cite{zheng2018magent, deka2021natural, fu2022concentration} 
or heterogeneous benchmarks consisting of only 2 agent types (e.g., most SMAC maps \cite{samvelyan2019starcraft})
to claim the effectiveness of their models.
Inhabiting a diverse world, 
we cannot rely solely on homogenous systems to efficiently solve sophisticated teamwork. 
Many real multiagent systems require diversified agents to participate to ensure the division of labor and lower costs. 
A typical example of this is found in colonies of social insects, such as honeybees and ants, 
where individual agents specialize in different tasks like foraging, nest construction and defense.
There are two primary challenges to address when applying MARL algorithms to heterogeneous multiagent tasks:

% 缺点1 non-stationary
Firstly,
non-stationarity \cite{hernandez2017survey} is one of the difficulties presented by heterogeneous multiagent problems.
Neglecting heterogeneous properties between agents can limit the performance of MARL models.
In heterogeneous multiagent tasks,
different kinds of agents vary in ability, quantity,
expected functionality, etc.
Notably, the number of each agent type is unbalanced in most scenarios.
E.g., in military simulations,
high-value aircraft are essential for obtaining battlefield advantage 
but are very few in number compared with ground agents.
This imparity is also common in the world of nature, 
e.g., in a hive, honey bee queens and workers differ significantly in number.
During an MARL training session,
agents that are relatively low in number suffer from the high variance in state-value estimation due to insufficient sampling,
which results in further non-stationary problems.

% TCA
% 缺点
Secondly,
heterogeneous systems contain more realistic problems that have not yet been studied 
in the context of decentralized automation systems.
Once a team of heterogeneous agents is trained and deployed, 
maintaining a distributed control system becomes an important issue and can cost considerable resources.
In the event of a system failure caused by a single type of agents, 
upgrading only the policies of the misbehaving agent type is a more reasonable and economical approach 
rather than upgrading the policies of the entire team. 
This requires agents to have the ability to cooperate with their heterogeneous teammates, 
even if they have different policy versions.
For example, 
if the flying policy of airborne agents is altered, 
it is important that other non-flying agents in the same heterogeneous system 
can acclimate to their teammates' policy shifts without requiring a series of chained policy updates.
Such capability is especially essential in automation systems that cannot enable Over-the-Air Programming (OTA)
due to reliability or security concerns (such as military robot systems),
or systems that cannot afford to suffer a full stop in long-term service 
(such as traffic and network devices).
Therefore, a method to iterate policies without breaking the compatibility 
between different heterogeneous agents is needed 
to reduce the cost of maintenance in actual applications of heterogeneous RL algorithms.
In this work, we summarize this compatibility issue as the policy version iteration problem.

% After a team of heterogeneous agents is trained and deployed,
% maintenance of a distributed control system is an important issue that can also cost considerable resources.
% When the system fails due to one of the agent types,
% it is more reasonably economical to only upgrade the policies of misbehaving agents rather than the whole team.
% This requires agents to have the capability to cooperate with heterogeneous teammates with different policy versions. 
% For instance, when the flying policy of airbone agents is changed,
% it is desirable that ground agent in the same heterogeneous system can adapt their teammates' policy shift
% without requiring a chained policy update.

% 贡献 check
In this paper, our contributions are as follows:
1. We propose Heterogeneous League Training (HLT), 
a general-purpose MARL algorithm for heterogeneous multiagent tasks.
2. We demonstrate the superior performance of HLT and the effectiveness of HLT in addressing the policy version iteration problem.
3. Based on HLT, we propose a method for evaluating
the difficulty of learning the roles undertaken by different agent types.
\section{Related Works}
% 异构强化学习 尚未有通用方法，及其原因是 ck
HMARL is not a new problem in MARL,
yet it has not yet been investigated extensively in the literature.
While there are works introducing model-based HMARL algorithms on specific domains, 
e.g., traffic control, medical information processing, and control system 
\cite{calvo2018heterogeneous, fard2022time},
a general-purpose heterogeneous reinforcement learning is absent as far as we're concerned.
On one hand,
tasks stressing their heterogeneous property are likely to have solid application backgrounds and concerns on realistic problems.
On the other hand,
existing simulation and training frameworks cannot support flexible task customizations
for heterogeneous systems.

% 异构强化学习环境 ck
While most RL simulation environments used in multiagent studies only consider 
homogeneous agent tasks \cite{deka2021natural, zheng2018magent, fu2022concentration},
benchmark environments that involve heterogeneous agent cooperation are also emerging.
For instance, Multi-Agent Particle Environment (MAPE) \cite{lowe2017multi}
is a simple but flexible multiagent environment,
designed for less complex tasks, such as predator-prey and speaker-listener tasks.
The SMAC benchmark environment proposed in \cite{samvelyan2019starcraft}
provides some examples of heterogeneous team combinations, such as 1c3s5z and MMM2.
Nevertheless, 
only a small proportion of the maps (such as MMM2) consider the supportive relationships within team agents.
% Moreover, despite there are tools to make SMAC maps, 
% SMAC still lacks customization ability for building new agents, 
% designing intra-team interactions or simulating more realistic tasks.

% 现有方法解决MARL的手段与区别ck
Existing MARL studies tend to address heterogeneous agent cooperation problems implicitly,
without distinguishing them especially from homogeneous ones \cite{rashid2018qmix, rashid2020weighted}.
Firstly,
all agents are provided with an identical action space, irrespective of their types.
If there are type-specific actions only available to certain agent types,
an action masking technique is used to handle these special cases.
Secondly,
agents are not explicitly modeled according to their types;
instead, 
agents need to discover their abilities and strengths by interacting with the environment and teammates.
This practice is successful in many task domains with weak agent ability distinction 
because agent experience can be shared regardless of agent types.
However, in complex heterogeneous problems, 
the agent policy cannot be generalized across different types,
preventing agents from benefiting from shared experiences with agents of different types.

% 列举一些MARL算法
In recent years, 
the majority of SOTA algorithms based on multiagent systems have been developed 
without explicit agent-type consideration.
In early models, such as IQL (Independently Q-Learning) \cite{watkins1989learning},
each agent learns its own policy independently.
According to \cite{lowe2017multi}, 
directly transferring single-agent approaches to multiagent environments results in a non-stationary problem \cite{hernandez2017survey}.
To address this issue, \cite{lowe2017multi} proposes a MADDPG algorithm, 
which is based on DDPG (Deep Deterministic Policy Gradient \cite{lillicrap2015continuous}), for stabilizing multiagent RL.
Moreover, a value decomposition algorithm VDN \cite{sunehag2017value} proposes a model based on Q-learning.
And the VDN model is further improved by Qmix \cite{rashid2018qmix} and Weighted Qmix \cite{rashid2020weighted},
which use hyper-networks \cite{ha2016hypernetworks} to decompose the task reward signals.
Nevertheless,
classic MARL methods ignore the discrepancies in the numbers and abilities of agents from different types,
which can negatively impact algorithm performance.

\begin{figure}[!b]
    \centering
    \includegraphics[width=0.9\linewidth]{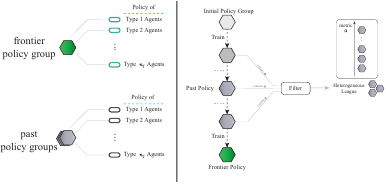}
    \caption{Heterogeneous League Training framework.}
    \label{fig:main-fig}
    \end{figure}

    % pay attention 
    % pay attention 
    % pay attention 
    % pay attention 
\section{Preliminaries}

\subsection{Heterogeneous MARL.}
% check
A heterogeneous MARL task can be described as a Dec-POMDP \cite{oliehoek2016concise} formulated by 
$\langle A, \varDelta, \mathcal{U}, \mathcal{S}, P_t, \mathcal{Z}, P_o, r , \gamma\rangle$,
where $A$ is the collection of agents and $N=|A|$ is the total number of agents.
$\mathcal{U}$ is a collection of actions within which each agent $a_i$ can choose.
$\mathcal{S}$ is the set of environment states and $P_t$ is the transition function.
$r$ is the team reward. $\gamma$ is the discount factor.
In a heterogeneous system, 
$\varDelta = \lbrace \delta_1, \dots, \delta_{n_j} \rbrace$ is the set of agent types and 
$n_j = |\varDelta|$ is the total number of types in the system.
Each agent $a_i$ is categorized into a type $\text{Type}(a_i) = \delta_j$.
Although the number of agents $N$ can be large,
the number of types $n_j$ is considerably smaller ($n_j \ll N$).
This is because real-world agents, typically robots, 
are usually manufactured using standardized production to reduce costs.
Furthermore, 
the numbers of different types of agents can vary significantly,
since it is common that high-value agents are fewer than 
cheap agents in a functional heterogeneous team.

\subsection{League Training in AlphaStar.}
% check
League training is first introduced in AlphaStar \cite{vinyals2019grandmaster}.
To enhance the initial capability of agents before optimizing them with reinforcement learning, 
AlphaStar takes advantage of data from human experts to train agents in a supervised approach,
obtaining a population of agents that possess diverse policies.
Agents are subsequently trained with reinforcement learning,
and intermittently duplicate themselves and freeze these duplications as past players.
The agents are trained against each other as well as past selves in history.
League training enables agents to learn adaptive and robust policies 
against the continually developing strategies of opponents. 
This characteristic also has significant value in heterogeneous MARL 
for addressing the policy version iteration problem and improving model performance.

% Unfortunately, league training cannot be applied to heterogeneous MARL 
% for the following reasons.
% First and most importantly,
% heterogeneous agents collaborate instead of competing in heterogeneous MARL.
% (If there are two or more adversarial teams, a unit entity is referred to as an agent only in its own team).
% Secondly,
% there will be no human data for supervised pre-training.
% Thirdly,
% League training used by \cite{vinyals2019grandmaster} consumes a huge amount of computational resources, 
% and it will be too expensive to be applied elsewhere.

%%%%%%%%%%%%%%%%%%%%%%%%%%%%%%%%%%%%%%%%%%%%%%%%%%%%%%
%%%%%%%%%%%%Heterogeneous League Training%%%%%%%%%%%%%
%%%%%%%%%%%%%%%%%%%%%%%%%%%%%%%%%%%%%%%%%%%%%%%%%%%%%%

\section{Heterogeneous League Training}

\begin{figure}[!t]
  \centering
  \includegraphics[width=\linewidth]{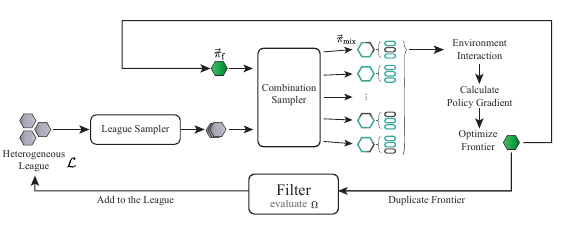}
  \caption{Policy optimization cycle.}
  \label{fig:optimization}
  \end{figure}

We propose a Heterogeneous League Training (HLT) algorithm,
an efficient and general-purpose 
multiagent RL algorithm for heterogeneous cooperative multiagent systems.

This section starts by stating the motivations and model features in Sec~\ref{sec:motivation}.
Next,
we briefly introduce the general framework of HLT in Sec~\ref{sec:framework}.
Then,
the details of heterogeneous league training are presented 
in Sec~\ref{sec:LeagueSampler} and Sec~\ref{sec:LeagueManagement}.
Finally,
as an essential structure to achieve adaptive cooperative behaviors,
the design of hyper-networks is illustrated in Sec~\ref{sec:hyper}.

\subsection{Motivation and Features.}
\label{sec:motivation}
HLT aims at two goals in heterogeneous multiagent RL:
\begin{itemize}
  \item Taking advantage of the feature of heterogeneous systems to facilitate cooperation.
  \item Addressing the policy version iteration problem for real-world automation system.
\end{itemize}

\subsubsection{Motivation} 

The following reasons inspire us to use the \textbf{league training} \cite{vinyals2019grandmaster} 
technique to achieve our goals.

% checked
(1) The numbers of different types of agents can vary considerably within a heterogeneous team.
Episode samples collected by the RL algorithm 
can provide low-variance experiences for types with numerous agents (T+). 
On the other hand, 
some types with comparably fewer agents (T-) receive high-variance experiences.
When T- agents fail to keep pace with T+ agents, the learning process becomes unstable.
Additionally, T+ agents tend to abandon their reliance on T- agents and 
instead develop uncooperative policies that exclude assistance from T- agents.
% Moreover, T+ agents tend to give up counting on T- agents and 
% turn to develop uncooperative policies that exclude the help of T-. 
By adopting the cooperative league training technique,
agents optimize their policies 
not only from the experiences of cooperating within a fixed team,
but also from the experiences of teaming up with foreign agents that possess distinct cooperation skills.
More specifically,
the cooperative league training technique can address the learning instability problem 
by encouraging agents to adopt different cooperation strategies 
depending on the unique traits of their teammates.

% checked
(2) Cooperative league training provides an environment where
agents are trained to work with a league of heterogeneous teammates 
equipped with diverse collaboration preferences.
By incorporating historical agent policies into the league,
agents can learn new cooperative behaviors 
without forgetting the way to collaborate with old versions of heterogeneous teammates.

\subsubsection{Distinctive features of HLT}

% checked
(1) 
In this paper, we propose a \textbf{cooperative} league training algorithm 
designed to facilitate cooperation among heterogeneous agents.
As a comparison, 
AlphaStar is an example of an \textbf{adversarial} multiagent league training algorithm \cite{vinyals2019grandmaster, han2020tstarbot}
where agents are trained to compete.

% checked
(2) The AlphaStar league model requires distributed servers with TPUs 
to run concurrent matches.
But our algorithm only needs a single machine with a single GPU 
to execute the entire training process.
To achieve this, 
we redesign the procedure of the league training based on 
the characteristic of heterogeneous systems.

% checked
(3) 
We propose a hyper-network structure, 
facilitating each agent with the capability to adapt its cooperative behavior
according to its type and the performance of its heterogeneous teammates.

\begin{figure*}[!t]
  \centering
  \includegraphics[width=0.57\linewidth]{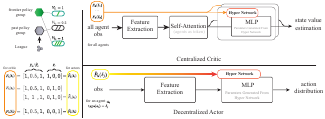}
  \caption{Hyper network for learning adaptive cooperative behavior.}
  \label{fig:hyper}
  \end{figure*}

\subsection{Overall Framework of HLT.}
\label{sec:framework}

\begin{algorithm}[h]
    \caption{HLT Policy Optimization}
    \label{alg:HLT}
    \begin{algorithmic}[1] %[1] enables line numbers
    \STATE Initialize a frontier policy group $\vec{\pi}_{\text{f}}$ and an empty league $\mathcal{L}$.
    \FOR {optimization step $M_s = 1,2,\dots$}
        \STATE Initialize an empty epsiode sample collection $\mathcal{D}$.
        \FOR {episode $M_e = 1,2,\dots, M_{em}$}
            \STATE Generate $C_{\text{sel}}$ with the league sampler from $\vec{\pi}_{\text{f}}$ and $\mathcal{L}$.
            \STATE Generate $\pi_{\text{mix}}$ with the combination sampler and $C_{\text{sel}}$.
            \STATE Run episode with the mixed policy group $\pi_{\text{mix}}$.
            \STATE Add epsiode samples to $\mathcal{D}$.
        \ENDFOR
        \STATE Calculate the policy gradient with $\mathcal{D}$.
        \STATE Update frontier policy networks $\vec{\pi}_{\text{f}}$.
        \IF {once every $M_{st}$ optimization steps}
            \STATE Evaluate $\vec{\pi}_{\text{f}}$ with metric $\Omega$
            \STATE Add $\vec{\pi}_{\text{f}}$ to $\mathcal{L}$ if $\vec{\pi}_{\text{f}}$ is accepted by the league filter.
        \ENDIF
        \STATE Update the critic network with $\mathcal{D}$.
    \ENDFOR
    \STATE End training.
    \end{algorithmic}
    \end{algorithm}

\subsubsection{Actors and Critic}
The HLT algorithm utilizes the actor-critic framework \cite{konda2000actor}
and follows the Centralized-Training-Decentralized-Execution (CTDE) paradigm \cite{lowe2017multi}.
As a prerequisite, 
all agents in the heterogeneous team are assumed to possess prior knowledge of their teammates' type identities.
The (actor) policy network parameters are shared among agents belonging to the same type.
However, agents of different types can freely choose between using shared parameters or independent networks.
We use a term \textbf{policy group} (denoted as $\vec{\pi}$) to represent all policies
of a heterogeneous team.
For generality,
policies for all agent types are represented by $\vec{\pi}$:
\begin{equation}
  \vec{\pi} = \lbrace \pi(\delta_1),\dots, \pi(\delta_{n_j}) \rbrace,
\end{equation}
where each agent type is represented by $ \delta_1, \dots, \delta_{n_j} $.
It is worth noting that each policy $\pi(\delta_j) \in \vec{\pi}$ 
is possessed by all agents that belong to type $\delta_j$ in the team.

To reduce the computational cost of training, 
we initialize and optimize only one dynamic learning policy group during league training.
This policy group $\vec{\pi}_{\text{f}}$ is referred to as the \textbf{frontier policy group}, 
or simply just \textbf{frontier} for clarity.

The HLT model employs a policy pool, 
also referred to as a \textbf{league} $\mathcal{L}$, 
to store historical policies visited by agents.
In HLT model, a policy pool, or a \textbf{league} $\mathcal{L}$, holds historical policies 
agents have visited.
This design bears a resemblance to the league structure of AlphaStar.
We only utilize one league structure in the algorithm.
As shown in Fig.\ref{fig:main-fig}, 
during the training process, 
a series of past policies are intermittently duplicated from the frontier and added to the league.
Once becoming independent and detached from the frontier,
the network parameters of the detached policies are frozen
so that we only need to compute policy gradient for frontier policy networks 
instead of for all policy networks in the league.
These frozen policies in the league are referred to as \textbf{past policy groups}.
Let $|\mathcal{L}|$ be the total number of policy groups in the league,
the league members (past policy groups) are denoted as:
\begin{equation}
  \mathcal{L}=\lbrace 
\vec{\pi}_{\text{p},1}, \dots, \vec{\pi}_{\text{p},|\mathcal{L}|}
 \rbrace,
\end{equation}

where $\vec{\pi}_{\text{p},1}, \dots, \vec{\pi}_{\text{p},|\mathcal{L}|}$ 
are the past policy groups in the league.

Additionally,
while there are different groups of policy networks (as actors) in HLT,
there is \textbf{only one critic network}
for action assessment during the entire training process.

\subsubsection{Policy Optimization Iteration}
%checked
As illustrated in Fig.~\ref{fig:main-fig},
the HLT algorithm begins by initializing random frontier policies $\vec{\pi}_{\text{f}}$ and an empty league.
Upon completion of each policy optimization iteration step, 
the new frontier policy group will be evaluated by a league filter,
determining whether it is eligible for inclusion as a new league member (Sec.~\ref{sec:LeagueManagement}).

%checked
Under the condition $|\mathcal{L}|=0$,
agents simply adopt the policies in the frontier $\vec{\pi}_{\text{f}}$.
As the training process progresses,
the league is incrementally populated with past policy groups.
Once $|\mathcal{L}|>0$,
mixed policies are used for policy optimization.
Complete details on policy mixing are provided in Sec.\ref{sec:LeagueSampler}.
% and collect episode samples for initial policy optimization.
% When the league is empty,

\subsection{Training with Mixed Policy Groups.}
\label{sec:LeagueSampler}
\label{sec:x}

%checked
As illustrated in Fig.~\ref{fig:optimization},
the frontier policy group is trained to cooperate not only with itself,
but also with past policies that visited previously during the learning process.
During training the heterogeneous team uses 
a mixed policy group denoted as $\vec{\pi}_{\text{mix}}$, 
which is a hybridization of frontier and past policies.
Two samplers are used to generate $\vec{\pi}_{\text{mix}}$ 
at the beginning of each training episode.

First,
a \textbf{league sampler} randomly generates a policy combination $C_{\text{sel}}$.
This combination is either frontier-past
$C_{\text{sel}}=(\vec{\pi}_{\text{f}}, \vec{\pi}_{\text{p},l})$, 
or frontier-frontier
$C_{\text{sel}}=(\vec{\pi}_{\text{f}} , \vec{\pi}_{\text{f}})$.
The balance between frontier-frontier and frontier-past combinations
is adjusted by a constant parameter $p_f \in \left[0,1\right)$:
\begin{equation}
\begin{array}{ll}
  &P\left[C_{\text{sel}}=\left( \vec{\pi}_{\text{f}}, \vec{\pi}_{\text{p},l} \right)    \mid p_f \right] = \frac{1}{(1-p_f)|\mathcal{L}|} \\
  &P\left[C_{\text{sel}}=\left( \vec{\pi}_{\text{f}}, \vec{\pi}_{\text{f}}          \right)    \mid p_f \right] = p_f,
\end{array}
\end{equation}
where $|\mathcal{L}|>0$ and $p_f<1$. 
When the league is empty ($|\mathcal{L}|=0$), 
the league sampler outputs only frontier-frontier combinations, 
namely $C_{\text{sel}}=( \vec{\pi}_{\text{f}}, \vec{\pi}_{\text{f}} )$.
For instance, when $p_f=0$, the sampler avoids pure frontier combinations 
and always samples one of the past policy groups from the league in $C_{\text{sel}}$.
Conversely, when $p_f \approx 1$,
the sampler tends to exclude the participant of past policies in the league.

Next, 
$C_{\text{sel}}$ is accepted by a \textbf{combination sampler},
which eventually determines the policy that each agent executes.
For clarity, we use $\vec{\pi}_\text{sel}$ and $C_{\text{sel}} = (\vec{\pi}_{\text{f}}, \vec{\pi}_\text{sel} )$ to represent 
the selected policy group and the combination, respectively.
As shown in Fig.~\ref{fig:optimization}, 
the combination sampler chooses one of the agent types to play the past policies.
Each type has an equal probability $P(\delta_j)$ of being selected,
regardless of the number of agents it possesses:
\begin{equation}
  P(\delta_j) = {1}/{|\varDelta|}, \forall \delta_j \in \varDelta.
\end{equation}

Finally, let $\delta_t$ be the agent type chosen by the combination sampler.
Then the mixed policy group $\vec{\pi}_{\text{mix}}$ that the team executes is represented by:
\begin{equation}
  \vec{\pi}_{\text{mix}} = \lbrace \pi_{\text{mix}}(\delta_1),\dots, \pi_{\text{mix}}(\delta_M) \rbrace,
\end{equation}
where
\begin{equation}
    \pi_{\text{mix}}(\delta_j) = \left\{\begin{array}{ll}
        \pi_f(\delta_j),            & \delta_j \neq  \delta_t \\
        \pi_{\text{sel}}(\delta_j), & \delta_j = \delta_t
    \end{array}\right., \forall \delta_j \in \varDelta.
\end{equation}

%checked
This mixed policy group $\vec{\pi}_{\text{mix}}$ is effective for only one episode.
And the league sampler and combination sampler can generate 
a batch of different $\vec{\pi}_{\text{mix}}$ to run in parallel.
This process is repeated to collect a batch of episodes for frontier policy optimization.
The policy gradient (w.r.t. the frontier policy parameters) is calculated from the collected episodes
and used to optimize each frontier policy of each agent type.

\begin{figure}[!t]
    \centering
    \begin{subfigure}[t]{0.48\linewidth}
    \centering
    \includegraphics[width=\linewidth,height=0.9\linewidth]{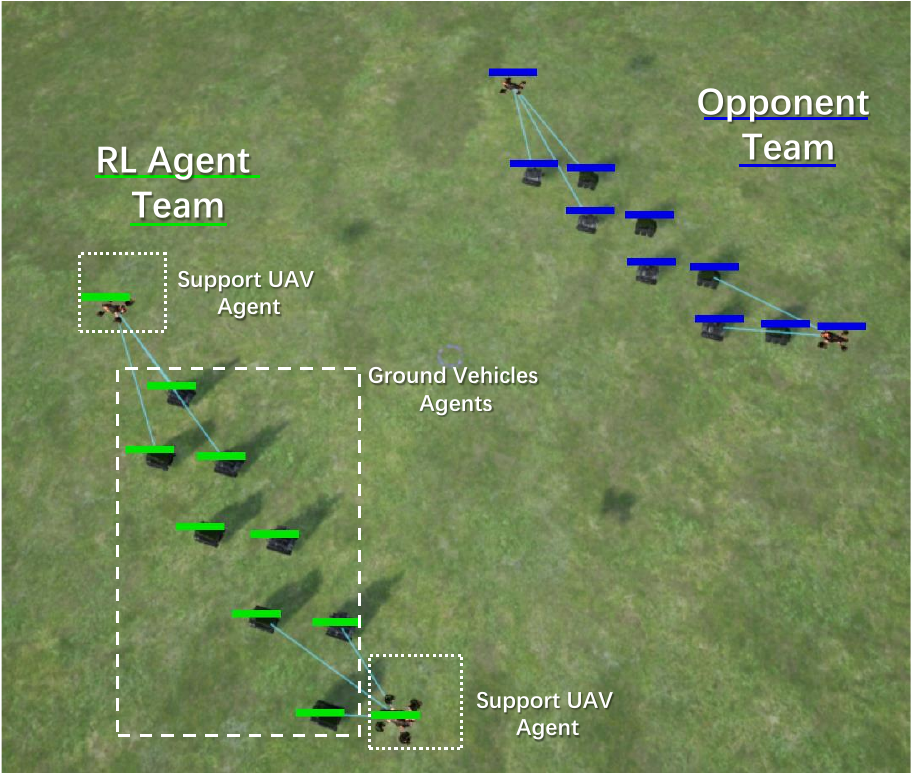}
    \caption{A map of UHMP with 2 support UAVs (T1) and 8 ground agents (T2 and T3) on each side.}
    \label{fig:uhmp-f1}
  \end{subfigure}
  \begin{subfigure}[t]{0.48\linewidth}
    \centering
    \includegraphics[width=\linewidth,height=0.9\linewidth]{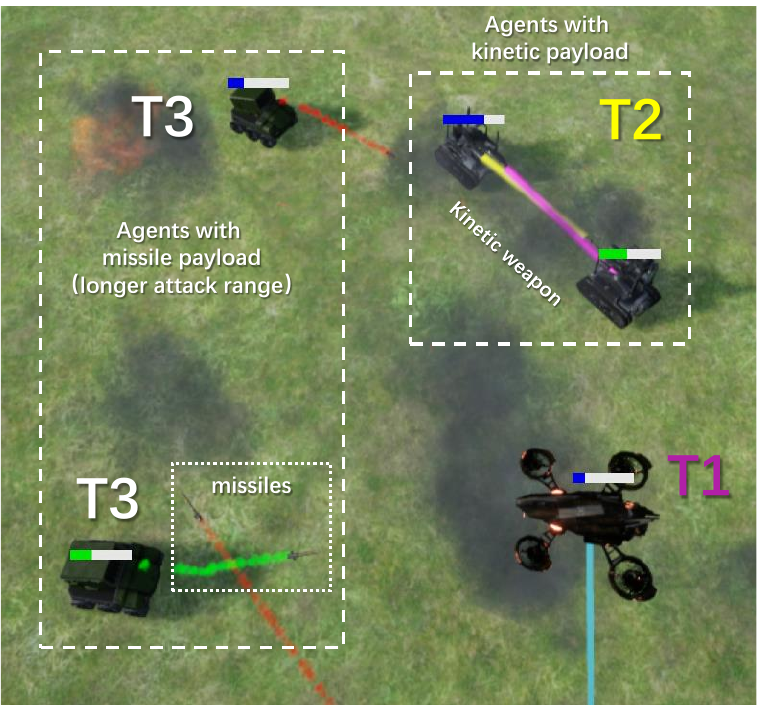}
    \caption{An illustration of three types of ground agents. Three agent types (T1, T2, T3) have distinctive abilities and responsibilities.}
    \label{fig:uhmp-f2}
  \end{subfigure}
\end{figure}

  % Supportive UAV agents (T1) are responsible for repairing nearby agents, 
  % equipped with missile-weapon payload but are vulnerable to opponent attacks.
  % Agents of the second type (T2) are ground units with missile weapon.
  % Agents of the third type (T3)  are ground units with kinetic weapon.

\subsection{League Member Management.}
\label{sec:LeagueManagement}
\label{sec:xx}

%checked
To be able to run on a single machine, the HLT model must control the size of the league. 
In comparison, 
researchers of AlphaStar utilize a large past agent pool, which costs a significant amount of memory and time.
The AlphaStar approach is not applicable to heterogeneous problems due to the significant cost of computational resources, 
which exceeds the limitations of most research facilities.
Consequently, we developed a league management component to decrease the size of the league. 
We had the intuition that the league should comprise policies demonstrating a variety of behaviors and unique cooperative abilities; 
hence, we created a league filter to manage and regulate its size. This filter is illustrated in Figure \ref{fig:optimization}.

%checked
In our HLT model, 
a metric $\Omega \in [0,1]$ is introduced to compare the similarity of policy groups.
This metric can be the success rate, normalized test reward 
(projecting the minimum reward and maximum reward to 0 and 1, respectively),
or any other evaluation scores that reflect the policy performance.

Let $|\mathcal{L}|_{\text{max}}$ be the maximum size of the league.
Whenever the size of the pool exceeds the $|\mathcal{L}|_{\text{max}}$ limitation,
the league filter deals with policy group candidates with the following procedures:
\begin{enumerate}
\item  Evaluate the new input policy group with metric $\Omega$.
\item  Add the policy group into the league, then sort all league policy groups in the league by their $\Omega$ scores.
\item  Locate the two policy groups with the closest $\Omega$ scores, then remove the newer group between them. 
\item  Repeat the last step until the size of the league is reduced to the maximum allowed size $|\mathcal{L}|_{\text{max}}$.
\end{enumerate}

For instance, after a new policy group is added temporarily to an already-full league in step 2), 
the league will have one extra member that needs to be removed:
\begin{equation}
\mathcal{L}=\lbrace \vec{\pi}_{\text{p},1}, \dots, \vec{\pi}_{\text{p},|\mathcal{L}|+1}  \rbrace.
\end{equation}

To locate a policy group to remove, 
past policy groups are evaluated by metric $\Omega$:
\begin{equation}
  \Omega(\mathcal{L}) 
= \lbrace \omega_1, \dots, \omega_{|\mathcal{L}|_{\text{max}}+1} \rbrace
= \left\lbrace   \Omega(\vec{\pi}_{\text{p},1}), \dots,    \Omega(\vec{\pi}_{\text{p},|\mathcal{L}|+1})     \right\rbrace,
\end{equation}
where $\omega_1, \dots, \omega_N, \omega_{|\mathcal{L}|_{\text{max}}+1}$ are evaluation scores.
Then the league is sorted according to these scores,
then rearrange $\mathcal{L}$ so that:
\begin{equation}
\omega_1 \le \dots \le \omega_N \le \omega_{|\mathcal{L}|_{\text{max}}+1}.
\end{equation}
Finally, find a pair of policy groups $(\vec{\pi}_{\text{p},k}, \vec{\pi}_{\text{p},k+1})$ with the closest evaluation metric:
\begin{equation}
|\omega_{k+1} - \omega_{k} | \le |\omega_{k'+1} - \omega_{k'} |, \quad \forall k' \in [1, |\mathcal{L}|_{\text{max}}].
\end{equation}
Considering the order of when group $k$ and group $k+1$ are added into the league,
we always remove the newer one between $\vec{\pi}_{\text{p},k}$ and $\vec{\pi}_{\text{p},k+1}$.
By removing the relatively newer group instead of the older one,
we can avoid frequent changes inside the league and prevent potential instability.
After removal, the size of the league is reduced back to $|\mathcal{L}|_{\text{max}}$.
% Then the order of when group $k$ and group $k+1$ are added into the pool is considered.
% We always remove the newer policy group within $\vec{\pi}_{\text{p},k}$ and $\vec{\pi}_{\text{p},k+1}$.

\subsection{Adaptive Hyper Network.}
\label{sec:hyper}
\label{sec:xxx}
% Cooperating with a variety of partners with different policies is challenging.
% Some partners are well-trained and willing to cooperate,
% while some partners using past policies frozen at not-trained states can only behave randomly.
% It is necessary to equip agents with a neural network structure 
% that is sensitive to the differences of heterogeneous partners. samples

Cooperating with a variety of partners with different policies is a challenging task.
Agents may encounter strong heterogeneous partners that are well-trained and cooperative. However,
they may also encounter weak heterogeneous partners that are equipped with immature policies sampled from the league.
As a result, 
it is necessary to equip agents with a neural network structure 
that is sensitive to the differences of heterogeneous partners.
In the HLT model, 
we use a \textbf{hyper-network} structure to address this problem.

As shown in Fig.~\ref{fig:hyper}, 
the hyper-network is a structure to generate network parameters for MLP layers in forward policy networks.
Internally, 
the hyper-network uses two layers of fully connected networks to produce MLP parameters,
using an agent-team information vector $F_h$ as input.

This agent-team information vector $F_h$ is determined by agent type $\delta_j$ 
as well as the mixed policy composition $\vec{\pi}_{\text{mix}}$. 
It is a concatenation of two parts:
\begin{equation}
F_h(\delta_j, \vec{\pi}_{\text{mix}}) = \operatorname{concat}\left[ \ 
    F_{v}(\delta_j, \vec{\pi}_{\text{mix}}) 
    , \ 
    F_{\delta}(\delta_j) \  
    \right],
\end{equation}
As it is shown in Fig.\ref{fig:hyper},
$F_{\delta}$ is a onehot binary vector to distinguish agent types. 
E.g., if $\delta_j=\delta_1$, then $F_{\delta}(\delta_j)=F_{\delta}(\delta_1)=[1, 0, 0, \dots]$.
And $F_{v}(\delta_j, \vec{\pi}_{\text{mix}})$ represents the combination of the mixed heterogeneous team 
by identifying the policy that each agent type executes. 
\begin{equation}
    F_{v}(\delta_j, \vec{\pi}_{\text{mix}})  = (V_{\delta_1},\dots,V_{\delta_{|\varDelta|}}),
\end{equation}
\begin{equation}
    V_{\delta_j}=\left\{\begin{array}{ll}
            1,                                      & \delta_j \neq  \delta_\text{sel} \\
            \Omega[\pi_{\text{sel}}(\delta_j)],     & \delta_j = \delta_\text{sel}
    \end{array}\right., \quad \forall \delta_j \in \varDelta,
\end{equation}
where $V_{\delta}$ distinguishes frontier policies (from past policies) with constant $ \Omega_\text{max}= 1$  for all $  \delta_j \neq  \delta_\text{sel} $. 
And past policies are represented with the metric score $\Omega$ of the policy group they are sampled from.
Using this method, $V_{\delta}$ can effectively illustrate the contrast between frontier policies and past policies,
since $\Omega[\pi_{\text{sel}}(\delta)] < 1$ unless agents achieve the maximum win rate or average reward (indicating the completion of training).

\begin{figure*}[!t]
  \centering
  \begin{subfigure}[t]{0.25\linewidth}
  \centering
  \includegraphics[width=\linewidth,height=0.6\linewidth]{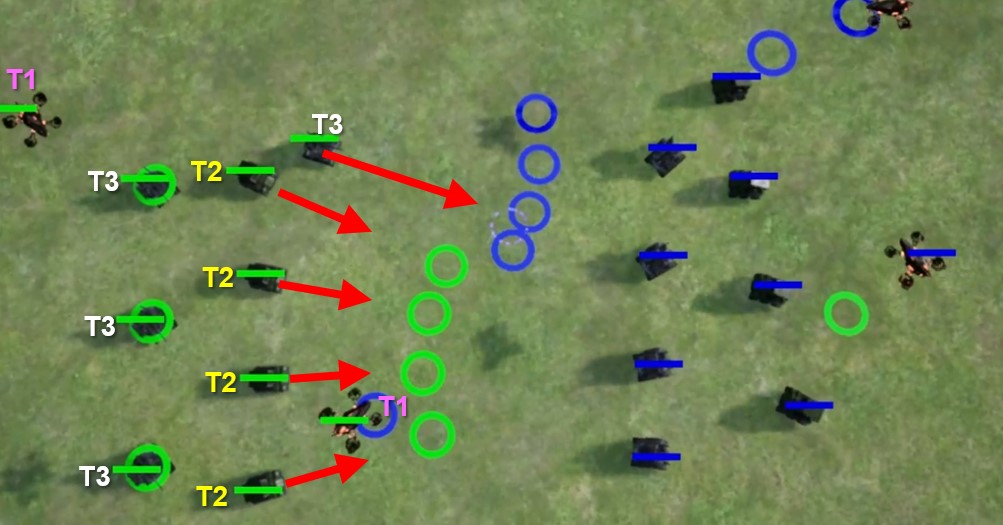}
  \caption{Step 4. T2 advancing.}
  \label{fig:dca1}
\end{subfigure}
\begin{subfigure}[t]{0.25\linewidth}
  \centering
  \includegraphics[width=\linewidth,height=0.6\linewidth]{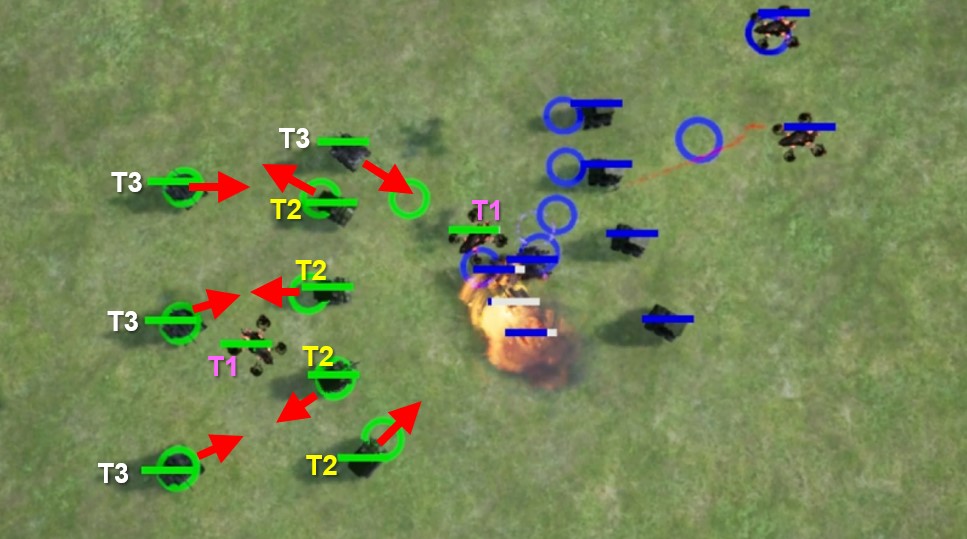}
  \caption{Step 8. T3 advancing.}
  \label{fig:dca2}
\end{subfigure}
\begin{subfigure}[t]{0.25\linewidth}
  \centering
  \includegraphics[width=\linewidth,height=0.6\linewidth]{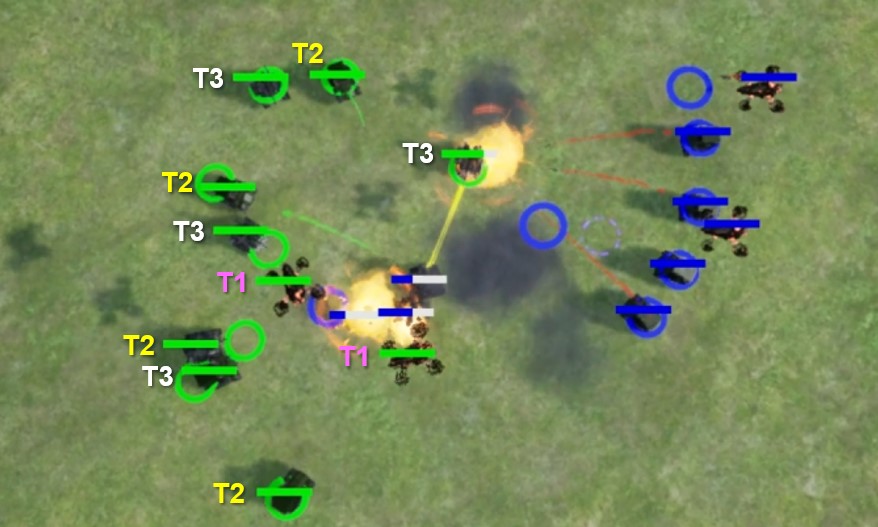}
  \caption{Step 11. Circle formation.}
  \label{fig:dca3}
\end{subfigure}
\caption{Cooperating with role division. HMARL algorithm controls the green team.
The moving direction of agents is represented by circles and red arrows nearby each agent in figures.}
\label{fig:uhmp}
\end{figure*}

Specially,
we also need to consider the hyper-network input for past policies.
As mentioned previously,
all past policies are duplications of the frontier policies at a certain stage.
Therefore, past policies are always trained under the circumstance 
where they observe themselves as the frontier. 
For consistency, 
past policies still need to observe themselves as frontiers (with $V_{\delta}=1$) even after they become a league member.
Concretely,
differing from $F_h$,
past policies eventually use $\hat{F}_h$ as the hyper-network input:
\begin{equation}
  \hat{F}_h(\delta_j, \vec{\pi}_{\text{mix}}) = \operatorname{concat}\left[ \ 
    [1,1,\dots,1]
    , \ 
    F_{\delta}(\delta_j) \  
    \right],
\end{equation}
Finally, the hyper-network produces a vector of parameters $\theta_h$:
\begin{equation}
\theta_h = \operatorname{HyperNet}[F_h(\delta_j, \vec{\pi}_{\text{mix}})],
\end{equation}
The output vector $\theta_h$ is split and then reshaped into the weight matrix and bias vector,
which function as parameters of policy MLP layers.
The hyper-network, together with the rest of the actor network, 
is trained in an unsupervised manner with the policy gradient.

% Two MLP layers are used as the structure of the hyper-network.
% For a heterogeneous team with agents $\lbrace a_1, \dots, a_{N} \rbrace$,
% we use a tuple $\langle \alpha,  \rangle$
% \begin{equation}

% \end{equation}

% describe the 
% aware of
% A hyper-network structure to solve the conflicts of cooperating with 
% different versions of agent

\section{Experiments}

\subsection{Heterogeneous Benchmark Environment.}
The HLT algorithm was implemented and tested within a heterogeneous simulation environment 
which we designed and named as Unreal-based Hybrid Multiagent Platform (UHMP).
As shown in Fig.\ref{fig:uhmp-f1} and Fig.\ref{fig:uhmp-f2},
we utilize a behavior-tree-based controller to play the opponent of our HLT learner.
Two teams share symmetrical configurations.
We use a tag \text{2u-4m-4k} to represent this configuration,
which indicates the type and number of heterogeneous agents in each team:

Each team comprises two supportive UAVs ($\delta_1$, \text{2u}), which have the ability to attack, repair allies, 
and mitigate damage from opponents, as well as eight ground units. 
Half of the ground units ($\delta_2$, \text{4m}) are armed with missile weapons that have a longer attack range, 
while the other half ($\delta_3$, \text{4k}) are equipped with kinetic weapons that have shorter attack ranges and are incapable of attacking opponent UAVs.
$\varDelta = \lbrace \delta_1, \delta_2, \delta_3 \rbrace$ and $|\varDelta| = 3$.

The functionalities of different agents in UHMP are also illustrated in Fig.\ref{fig:uhmp-f2} and Fig.\ref{fig:uhmp-f2}.
Due to the ability difference,
agents belonging to one type have to learn policies that are, to an extent, unshareable to other heterogeneous types.
Furthermore,
due to the imparity of quantity between different types,
the learning difficulty of different agent types diverges from one another.

\subsection{Experimental Setup.}

We run our experiments on a single Linux server with Nvidia GPUs (RTX 8000).
Only one GPU was used for each experiment run.
We adopt dual-clip PPO \cite{ye2020mastering}, Adam optimizers and 
policy resonance \cite{fu2022solving} techniques to improve training efficiency.
To control the involvement of the heterogeneous league,
we set $p_f=0.1$ for the league sampler by default,
so that agents execute mixed policies 
with the participant of both frontier and past policies in 90\% of the episodes.
We set $|\mathcal{L}|_\text{max} = 5$ because running a large number of past policy groups concurrently 
can significantly slow down the model calculation.
For each optimization step,
we use samples collected from $M_{em}=128$ episodes to calculate the policy gradient.
We frequently evaluate the frontier policy group with the winning rate as the metric $\Omega$,
and we average the results of 160 episodes in every evaluation for precision.
The intermittent evaluation takes place once every $M_{st}=1600$ training episodes.
All sets of experiments were repeated four times with different random seeds.

\subsection{Performance Evaluation of HLT.}
To test the effectiveness of the proposed HLT model,
we compare the proposed HLT method with 
FT-Qmix (Fine-Tuned Qmix) \cite{hu2021rethinking}.

While individual observations alone suffice the requirements of our HLT model,
FT-Qmix requires an extra state information vector during the centralized training stage.
Determining how the additional states are obtained,
we use the '-D' suffix to represent a model trained 
in a Delicately designed state space,
or no suffix to represent a model trained in an untuned state space (or not using state information at all).

\begin{figure}[!t]
  \centering
  \begin{subfigure}[t]{0.48\linewidth}
  \centering
  \includegraphics[width=\linewidth,height=0.8\linewidth]{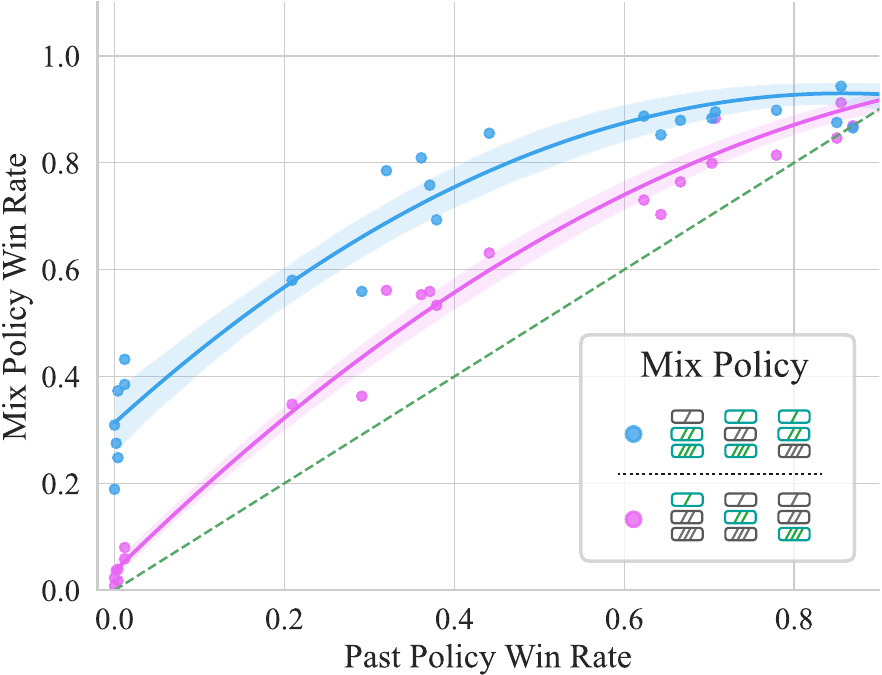}
  \caption{Compatibility between eventual frontier policies and past policies.}
  \label{fig:win_rate_absolute_improvement_both}
\end{subfigure}
\begin{subfigure}[t]{0.48\linewidth}
  \centering
  \includegraphics[width=\linewidth,height=0.8\linewidth]{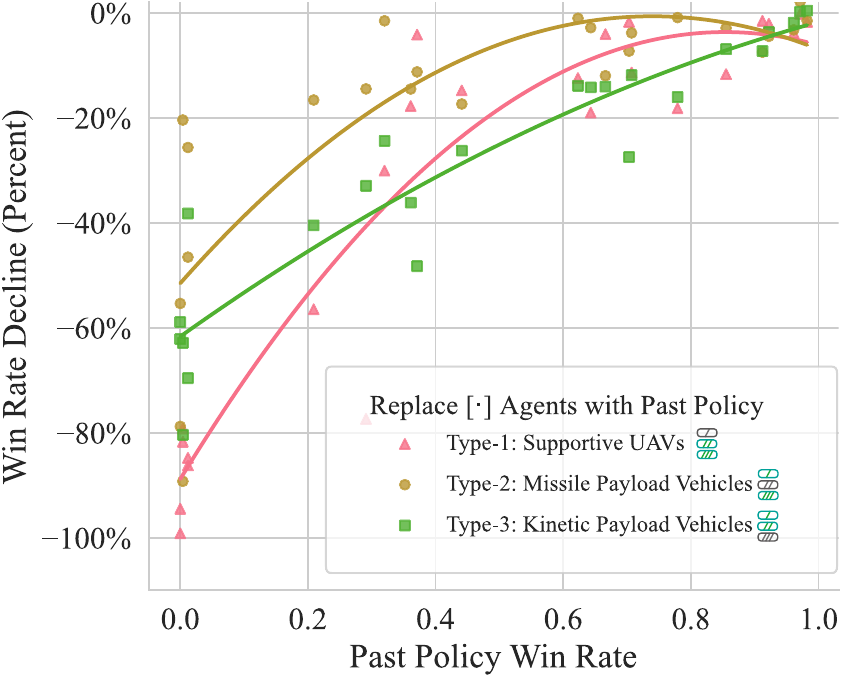}
  \caption{Heterogeneous role learning process when mixing the frontier and the past policies. (Frontier exclusive.)}
  \label{fig:ReplaceTest1}
\end{subfigure}
\caption{Compatibility studies.}
\end{figure}

% \begin{figure}[!t]
%   \centering
%   \includegraphics[width=0.5\linewidth]{img/win_rate_absolute_improvement_both_addicon.pdf}
%   \caption{Compatibility between eventual frontier policies and past policies.}
%   \label{fig:win_rate_absolute_improvement_both}
% \end{figure}

% \begin{figure}[!t]
%   \centering
%   \includegraphics[width=0.5\linewidth]{img/ReplaceTest1_addIcon_final.pdf}
%   \caption{Revealing heterogeneous role learning process by mixing the frontier and the past policies. (Frontier exclusive.)}
%   \label{fig:ReplaceTest1}
% \end{figure}

% We investigate how
% We examine whether both chosen motivational indices are essential.
% (a) We assess how the concentration parameter dc influences the module. 
% (b) We investigate whether the extra
% self-attention layer shown in Fig. 1 provides improvement.
% (c) We examine whether reducing our model from DualConcNet to Single-ConcNet causes a performance decay.
% (d) We investigate whether Conc-4Hist model can utilize
% history observations under extreme interference.

\subsection{The Effectiveness Against Policy Version Iteration.}
\label{PolicyIteration}

% we can then test whether a fully trained policy group (the eventual frontier policy group) can promote the win rate of past policies.
After completing the training stage, 
we evaluate the compatibility of the fully trained policy group (namely the eventual frontier policy group). 
To determine their affinity, 
we assess the performance of mixture policies comprising both past policies and the eventual frontier policy.
In this test,
we once again leverage the random combination sampler introduced previously to carry out the following tests:

(\textbf{1}) In each episode, randomly select one type of agent that executes the past policies 
    while the types of agents execute the frontier policies. (Frontier exclusive test.)

(\textbf{2}) In each episode, randomly select one type of agent that executes the frontier policies, 
    while the types of agents execute the past policies. (Frontier inclusive test.)
% For each episode, we randomly select one type of agent that executes the past policies 
% while the types of agents execute the frontier policies. % (Frontier Exclusive.)

In the first test, 
most agents (and agent types) are controlled by the frontier policies.
With the frontier policy group being the majority in the team, 
we can assess how the frontier works with the minority of agents 
controlled by past policies with various cooperation capabilities.
In the second evaluation, 
we assessed how well the trained frontier policies could adapt in a scenario where most agents follow past policies. 
In this case, frontier policies controlled a minority of agents and agent types in the team, 
which challenged their adaptability in the absence of a dominant position.

% The Differences Between Heterogeneous Roles
\subsection{Bottleneck Investigation in Heterogeneous Team.}
An agent type can become the bottleneck of a heterogeneous team if its role is relatively more difficult to learn. 
Next, we demonstrate that 
the HLT model is capable of statistically assessing the learning process of different agent types, 
and revealing the bottleneck agent type during different learning phases.

Similar to the previous policy version iteration evaluation,
we again mix the eventual frontier policies with past policies.
A small difference is that the agent type to execute the past policy 
is specially assigned rather than randomly chosen.
This experiment assesses the difficulty of learning diverse roles, 
and furthermore provides a new perspective on how cooperative heterogeneous joint policies are formed from scratch.

% This experiment provides us with a method to assess the difficulty of learning each heterogeneous role,
% and furthermore offers a new perspective for analyzing 
% how a cooperative heterogeneous joint policy is formed from scratches.

% except that we seperate the result according to 
% which type of agents is chosen to execute the past policy.

% the difficulty of learning policies for different types 

% In the policy iteration test, 
% we sample the type combination randomly and 
% only focus on the frontier's adaptive ability as a group.

% To
% We investigate the performance differences 
% as agents 

\section{Results.}

\subsection{Main Results.}
We compared our HLT method with FT-Qmix on two main metrics, 
namely win rate and averaged test reward.
Each score is averaged over 160 test episodes, 
and then averaged again over 4 independent experiments launched by different random seeds.
According to Table.1,
HLT has the best performance among all models,
despite the fact that HLT does not use any state information.
% FT-Qmix also has a relatively good performance 
% when trained by a delicately designed global state.

%%%%%%%%%%%%%%%%%%%%%%%%%%%%%%%%%%%%%%%%%%%%%%%%%%
%%%%%%%%%%%%%%%%%%%%%%%%%%%%%%%%%%%%%%%%%%%%%%%%%%
%%%%%%%%%%%%%%%%%%%%%%%%%%%%%%%%%%%%%%%%%%%%%%%%%%
%%%%%%%%%%%%%%%%%%%%%%%%%%%%%%%%%%%%%%%%%%%%%%%%%%
\subsection{Heterogeneous Cooperation Policy.}
Our observations of the eventual frontier policies trained by HLT 
indicate a distinct division of agent roles between the different agent types.
Figure \ref{fig:uhmp} depicts the situation 
where green RL agents are poised to make contact with the enemy.
The UAVs, or T1 agents, 
take advantage of their agility 
by maneuvering back and forth to attack opponents and provide support to ground teammates.
In contrast, T2 agents have a longer attack range than T3 but employ a hit-and-run policy. 
They advance in front of T3 agents when enemies are far away 
but quickly retreat behind T3 when danger is imminent.
T3 agents are skilled in short-range combat. 
However, as part of a heterogeneous team, 
they adopt a more restrained policy and remain stationary at the beginning of the simulation.
To obtain advantages when opponents approach, 
T3 agents join forces with T2 agents to form a semi-circle-shaped formation.

% We observed that the eventual frontier policies trained by HLT 
% display a clear division of agent roles across different agent types.
% Fig.\ref{fig:uhmp} demonstrates a scenario where 
% green RL agents are preparing to contact the enemy.
% T1 agents (UAVs) take advantage of their agility 
% by moving back and forth 
% to attack enemies and simultaneously support ground teammates.
% T2 agents have a longer attack range than T3,
% but they learned a hit-and-run policy,
% moving ahead of T3 agents when opponents are far but 
% immediately retreating behind T3 agents when facing imminent threats.
% The strength of T3 agents is short-range combat.
% But being in a heterogeneous team, 
% they learn a rather sedate policy to stay still when the simulation begins.
% To obtain advantages when opponents approach,
% T3 agents form a semi-circle-shaped formation along with T2 agents.

% \begin{table}[h]
%   \centering
%   \renewcommand\arraystretch{1.6} %修改行边距
%   \tabcolsep=0.16cm %修改列边距
%   \caption{SOME PARAMETERS SETTING FOR PROPOSED METHOD}
%   \begin{tabular}{c|c|c|c|c|c} \hline \hline % ???|c|?????????????????
%    \textbf{Symbol} & \textbf{Value} & \textbf{Symbol} & \textbf{Value} & \textbf{Symbol} & \textbf{Value} \\ \hline
%    ${\sigma _1}$ & $5$     & $k_j$     & $0.4$   & $R_c$  & $30$ \\  \hline 
%    ${\sigma _2}$ & $25$    & $k_b$     & $0.4$   & $k_c$  & $10$ \\ \hline
%    $Q$           & $100$   & $k_v$     & $0.3$   & $p_a$  & $0.15$ \\ \hline
%    $\varepsilon$ & $0.1$   & $r_a$     & $10$    & $p_t$  & $0.3$ \\ \hline \hline
%   \end{tabular}
%   \label{parameters}
%  \end{table}

\begin{table}[!t]
  % \caption{}
  \label{table:performance}
  \begin{subtable}[h]{0.45\textwidth}
     \centering
     \caption{Table 1. Performance comparison between methods.}
     \begin{tabular}{ccc}
      \toprule
      {Algorithm} 
                                      & {Best Test Win Rate}  
                                                              & {Best Test Reward}         \\
      \midrule
      {HLT         }  
                                      & \textbf{98.09\%}              
                                                              & \textbf{1.782}      \\
      {FT-Qmix} 
                                      & {87.98\%}              
                                                              & {1.452}   \\
      {FT-Qmix-D}              
                                      & {91.79\%}              
                                                              & {1.547}   \\
      \bottomrule
  \end{tabular}
  \end{subtable}

\end{table}

%%%%%%%%%%%%%%%%%%%%%%%%%%%%%%%%%%%%%%%%%%%%%%%%%%
%%%%%%%%%%%%%%%%%%%%%%%%%%%%%%%%%%%%%%%%%%%%%%%%%%
%%%%%%%%%%%%%%%%%%%%%%%%%%%%%%%%%%%%%%%%%%%%%%%%%%
%%%%%%%%%%%%%%%%%%%%%%%%%%%%%%%%%%%%%%%%%%%%%%%%%%
\subsection{Policy Compatibility.}

Fig.\ref{fig:win_rate_absolute_improvement_both}
demonstrates the win rate improvement when
the frontier policy group interferes with the past policy groups.
The past policy groups have different levels of cooperative ability measured using their $\Omega$ metric (policy group win rate).
In this experiment, we use past policy groups satisfying $\Omega < 0.90$,
which are distinguishable from the frontier policy group with $\Omega \approx 1$.
We leverage two tests described in Sec~\ref{PolicyIteration}.
In Fig.\ref{fig:win_rate_absolute_improvement_both},
frontier-exclusive test results are represented by blue samples,
and frontier-inclusive test results are shown with red samples.
A significant improvement can be observed in the policy group 
with $\Omega \in (0.2, 0.6)$ in both tests.
The participants of the eventual frontier policy promote
the win rate notably above the green $\Omega$ baseline,
suggesting an affinity can be established between the frontier and past policies.
When $\Omega > 0.7$,
the room for improvement is limited.
Nonetheless, frontier policies can still provide compatibility
without raising conflicts that ruin the cooperation win rate in those experiments.

% \begin{figure}[h]
%   \centering
%   \includegraphics[width=0.6\linewidth]{img/win_rate_improvement_both.pdf}
%   \caption{Compatibility between finalized policies and past policies. 
%   Improvement with the participant of eventual frontier policies.
%   }
%   \label{fig:PolicyIteration}
% \end{figure}

% \begin{figure}[h]
%   \centering
%   \includegraphics[width=0.6\linewidth]{img/win_rate_improvement_both.pdf}
%   \caption{A wrong images.}
%   \label{fig:win_rate_improvement_both}
% \end{figure}

%%%%%%%%%%%%%%%%%%%%%%%%%%%%%%%%%%%%%%%%%%%%%%%%%%
%%%%%%%%%%%%%%%%%%%%%%%%%%%%%%%%%%%%%%%%%%%%%%%%%%
%%%%%%%%%%%%%%%%%%%%%%%%%%%%%%%%%%%%%%%%%%%%%%%%%%
%%%%%%%%%%%%%%%%%%%%%%%%%%%%%%%%%%%%%%%%%%%%%%%%%%
\subsection{Heterogeneous Role Learning.}

% \begin{figure}[h]
%   \centering
%   \includegraphics[width=0.6\linewidth]{img/ReplaceTest2.pdf}
%   \caption{Revealing heterogeneous role learning process 
%   by mixing the frontier and the past policies. (Frontier inclusive.)}
%   \label{fig:ReplaceTest2}
% \end{figure}

% All types of agents are essential in heterogeneous tasks.
% Nevertheless,
% different agent types vary not only in their ability and responsibility,
% but also in their difficulty to master and learn.

% However,
% this result changes as the training procceds,
% T1 agents catch up fast with other agents 
% when the policy group reaches $\Omega \approx 0.4$ during the learning process.
The HLT model enables us to evaluate and monitor the learning progress of individual agent types.
Fig.\ref{fig:ReplaceTest1} reveals the win rate decline 
when agents of a specific type switch to execute past policies instead of the eventual frontier.
For example,
we can compare the consequence of replacing the policy of T1, T2 and T3 with past policies 
with $\Omega \approx 0$ (bad random policies) respectively.
The worst consequence occurs when we replace the policy of T1 (UAVs),
which causes the most serious consequences (reducing win rate by 90\% on average).
In comparison, replacing the policy of T2 had a relatively smaller impact.
This suggests that well-functioning T1 agents 
are most indispensable in the frontier policies compared with the other two agent types.
However, this result changes as the training proceeds. 
T1 agents quickly catch up with other agents at $\Omega \approx 0.4$ during the learning process.

On the other hand,
T3 agents have the slowest $\Omega$ metric ascent among all three agent types,
and usually bring the most significant win rate decline when team $\Omega > 0.4$.
This indicates that T3 agents are typically bottlenecks in the training process.

% HLT和其他方法的胜率对比

% HLT和不使用HLT前后，策略兼容性对比

% HLT策略兼容性的进一步研究，定量化地展示每一个类别智能体对任务的

\section{Conclusions}
This work presents a Heterogeneous League Training algorithm 
for addressing general heterogeneous multiagent challenges.
HLT introduces a lightweight league component that is composed of diverse agent policies 
accumulated at various training stages,
thereby promoting the robustness of agent policies by training agents
with heterogeneous teammates exhibiting different cooperation skills.

We demonstrate that the HLT algorithm performs better 
compared to other methods in cooperative heterogeneous tasks. 
Additional policy compatibility experiments show that 
policies trained by HTL
are capable of cooperating with immature past policies and promoting their win rate.
Moreover, HLT can statistically assess the degree of learning difficulty 
for different roles in a heterogeneous team, 
thereby revealing the vulnerabilities of the cooperative system more effectively.
% Moreover,
% HLT can statistically evaluate the difficulty of learning 
% different roles in a heterogeneous team,
% making it easier to identify the weakness of the heterogeneous cooperative system.

\bibliography{cite.bib}

\begin{thebibliography}{22}
\providecommand{\natexlab}[1]{#1}
\providecommand{\url}[1]{\texttt{#1}}
\providecommand{\urlprefix}{URL }
\expandafter\ifx\csname urlstyle\endcsname\relax
  \providecommand{\doi}[1]{doi:\discretionary{}{}{}#1}\else
  \providecommand{\doi}{doi:\discretionary{}{}{}\begingroup
  \urlstyle{rm}\Url}\fi

\bibitem[{Calvo and Dusparic(2018)}]{calvo2018heterogeneous}
Calvo, J.A. and Dusparic, I. (2018).
\newblock Heterogeneous multi-agent deep reinforcement learning for traffic
  lights control.
\newblock In \emph{AICS}, 2--13.

\bibitem[{Deka and Sycara(2021)}]{deka2021natural}
Deka, A. and Sycara, K. (2021).
\newblock Natural emergence of heterogeneous strategies in artificially
  intelligent competitive teams.
\newblock In \emph{International Conference on Swarm Intelligence}, 13--25.
  Springer.

\bibitem[{Fard and Selmic(2022)}]{fard2022time}
Fard, E. and Selmic, R.R. (2022).
\newblock Time-delayed data transmission in heterogeneous multi-agent deep
  reinforcement learning system.
\newblock In \emph{2022 30th Mediterranean Conference on Control and Automation
  (MED)}, 636--642. IEEE.

\bibitem[{Fu et~al.(2022{\natexlab{a}})Fu, Qiu, Yi, Pu, Ai, and
  Yuan}]{fu2022solving}
Fu, Q., Qiu, T., Yi, J., Pu, Z., Ai, X., and Yuan, W. (2022{\natexlab{a}}).
\newblock Solving the diffusion of responsibility problem in multiagent
  reinforcement learning with a policy resonance approach.
\newblock \emph{arXiv preprint arXiv:2208.07753}.

\bibitem[{Fu et~al.(2022{\natexlab{b}})Fu, Qiu, Yi, Pu, and
  Wu}]{fu2022concentration}
Fu, Q., Qiu, T., Yi, J., Pu, Z., and Wu, S. (2022{\natexlab{b}}).
\newblock Concentration network for reinforcement learning of large-scale
  multi-agent systems.
\newblock \emph{Proceedings of the AAAI Conference on Artificial Intelligence},
  36(9), 9341--9349.
\newblock \doi{10.1609/aaai.v36i9.21165}.

\bibitem[{Ha et~al.(2016)Ha, Dai, and Le}]{ha2016hypernetworks}
Ha, D., Dai, A., and Le, Q.V. (2016).
\newblock Hypernetworks.
\newblock \emph{arXiv preprint arXiv:1609.09106}.

\bibitem[{Han et~al.(2020)Han, Xiong, Sun, Sun, Fang, Guo, Chen, Shi, Yu, Wu
  et~al.}]{han2020tstarbot}
Han, L., Xiong, J., Sun, P., Sun, X., Fang, M., Guo, Q., Chen, Q., Shi, T., Yu,
  H., Wu, X., et~al. (2020).
\newblock Tstarbot-x: An open-sourced and comprehensive study for efficient
  league training in starcraft ii full game.
\newblock \emph{arXiv preprint arXiv:2011.13729}.

\bibitem[{Hernandez-Leal et~al.(2017)Hernandez-Leal, Kaisers, Baarslag, and
  De~Cote}]{hernandez2017survey}
Hernandez-Leal, P., Kaisers, M., Baarslag, T., and De~Cote, E.M. (2017).
\newblock A survey of learning in multiagent environments: Dealing with
  non-stationarity.
\newblock \emph{arXiv preprint arXiv:1707.09183}.

\bibitem[{Hu et~al.(2021)Hu, Jiang, Harding, Wu, and wei
  Liao}]{hu2021rethinking}
Hu, J., Jiang, S., Harding, S.A., Wu, H., and wei Liao, S. (2021).
\newblock Rethinking the implementation tricks and monotonicity constraint in
  cooperative multi-agent reinforcement learning.

\bibitem[{Konda and Tsitsiklis(2000)}]{konda2000actor}
Konda, V.R. and Tsitsiklis, J.N. (2000).
\newblock Actor-critic algorithms.
\newblock In \emph{Advances in Neural Information Processing Systems},
  1008--1014.

\bibitem[{Kr{\"{o}}se(1995)}]{watkins1989learning}
Kr{\"{o}}se, B.J.A. (1995).
\newblock Learning from delayed rewards.
\newblock \emph{Robotics Auton. Syst.}, 15(4), 233--235.
\newblock \doi{10.1016/0921-8890(95)00026-C}.

\bibitem[{Lillicrap et~al.(2015)Lillicrap, Hunt, Pritzel, Heess, Erez, Tassa,
  Silver, and Wierstra}]{lillicrap2015continuous}
Lillicrap, T.P., Hunt, J.J., Pritzel, A., Heess, N., Erez, T., Tassa, Y.,
  Silver, D., and Wierstra, D. (2015).
\newblock Continuous control with deep reinforcement learning.
\newblock \emph{arXiv preprint arXiv:1509.02971}.

\bibitem[{Lowe et~al.(2017)Lowe, Wu, Tamar, Harb, Abbeel, and
  Mordatch}]{lowe2017multi}
Lowe, R., Wu, Y., Tamar, A., Harb, J., Abbeel, P., and Mordatch, I. (2017).
\newblock Multi-agent actor-critic for mixed cooperative-competitive
  environments.
\newblock \emph{arXiv preprint arXiv:1706.02275}, 6382--6393.

\bibitem[{Oliehoek and Amato(2016)}]{oliehoek2016concise}
Oliehoek, F.A. and Amato, C. (2016).
\newblock \emph{A concise introduction to decentralized POMDPs}.
\newblock Springer.

\bibitem[{Rashid et~al.(2020)Rashid, Farquhar, Peng, and
  Whiteson}]{rashid2020weighted}
Rashid, T., Farquhar, G., Peng, B., and Whiteson, S. (2020).
\newblock Weighted qmix: Expanding monotonic value function factorisation.
\newblock \emph{arXiv e-prints}, arXiv--2006.

\bibitem[{Rashid et~al.(2018)Rashid, Samvelyan, Schroeder, Farquhar, Foerster,
  and Whiteson}]{rashid2018qmix}
Rashid, T., Samvelyan, M., Schroeder, C., Farquhar, G., Foerster, J., and
  Whiteson, S. (2018).
\newblock Qmix: Monotonic value function factorisation for deep multi-agent
  reinforcement learning.
\newblock In \emph{International Conference on Machine Learning}, 4295--4304.
  PMLR.

\bibitem[{Samvelyan et~al.(2019)Samvelyan, Rashid, De~Witt, Farquhar, Nardelli,
  Rudner, Hung, Torr, Foerster, and Whiteson}]{samvelyan2019starcraft}
Samvelyan, M., Rashid, T., De~Witt, C.S., Farquhar, G., Nardelli, N., Rudner,
  T.G., Hung, C.M., Torr, P.H., Foerster, J., and Whiteson, S. (2019).
\newblock The starcraft multi-agent challenge.
\newblock \emph{arXiv preprint arXiv:1902.04043}, 2186--2188.

\bibitem[{Sunehag et~al.(2017)Sunehag, Lever, Gruslys, Czarnecki, Zambaldi,
  Jaderberg, Lanctot, Sonnerat, Leibo, Tuyls et~al.}]{sunehag2017value}
Sunehag, P., Lever, G., Gruslys, A., Czarnecki, W.M., Zambaldi, V., Jaderberg,
  M., Lanctot, M., Sonnerat, N., Leibo, J.Z., Tuyls, K., et~al. (2017).
\newblock Value-decomposition networks for cooperative multi-agent learning.
\newblock \emph{arXiv preprint arXiv:1706.05296}.

\bibitem[{Vinyals et~al.(2019)Vinyals, Babuschkin, Czarnecki, Mathieu, Dudzik,
  Chung, Choi, Powell, Ewalds, Georgiev et~al.}]{vinyals2019grandmaster}
Vinyals, O., Babuschkin, I., Czarnecki, W.M., Mathieu, M., Dudzik, A., Chung,
  J., Choi, D.H., Powell, R., Ewalds, T., Georgiev, P., et~al. (2019).
\newblock Grandmaster level in starcraft ii using multi-agent reinforcement
  learning.
\newblock \emph{Nature}, 575(7782), 350--354.

\bibitem[{Ye et~al.(2020)Ye, Liu, Sun, Shi, Zhao, Wu, Yu, Yang, Wu, Guo
  et~al.}]{ye2020mastering}
Ye, D., Liu, Z., Sun, M., Shi, B., Zhao, P., Wu, H., Yu, H., Yang, S., Wu, X.,
  Guo, Q., et~al. (2020).
\newblock Mastering complex control in moba games with deep reinforcement
  learning.
\newblock In \emph{Proceedings of the AAAI Conference on Artificial
  Intelligence}, volume~34, 6672--6679.

\bibitem[{Zhao et~al.(2019)Zhao, Zong, Tian, Zhang, and You}]{zhao2019fast}
Zhao, X., Zong, Q., Tian, B., Zhang, B., and You, M. (2019).
\newblock Fast task allocation for heterogeneous unmanned aerial vehicles
  through reinforcement learning.
\newblock \emph{Aerospace Science and Technology}, 92, 588--594.
\newblock \doi{10.1016/j.ast.2019.06.024}.

\bibitem[{Zheng et~al.(2018)Zheng, Yang, Cai, Zhou, Zhang, Wang, and
  Yu}]{zheng2018magent}
Zheng, L., Yang, J., Cai, H., Zhou, M., Zhang, W., Wang, J., and Yu, Y. (2018).
\newblock Magent: A many-agent reinforcement learning platform for artificial
  collective intelligence.
\newblock In \emph{Proceedings of the AAAI Conference on Artificial
  Intelligence}, volume~32.

\end{thebibliography}

\end{document}